\documentclass{article}

\usepackage{microtype}
\usepackage{graphicx}
\usepackage{subfigure}
\usepackage{booktabs} 

\usepackage{hyperref}

\usepackage[accepted]{icml2023}

\usepackage{amssymb}
\usepackage{mathtools}
\usepackage{amsthm}
\usepackage{amsmath}
\usepackage{booktabs}
\usepackage[utf8]{inputenc}
\usepackage{nicematrix}

\usepackage{algorithm} 
\usepackage{algorithmic}  
\usepackage[algo2e]{algorithm2e} 

\usepackage[capitalize,noabbrev]{cleveref}

\theoremstyle{plain}

\theoremstyle{definition}

\theoremstyle{remark}

\usepackage[textsize=tiny]{todonotes}

\icmltitlerunning{Submission and Formatting Instructions for ICML 2023}

\begin{document}

\twocolumn[
\icmltitle{Learning To Explore With Predictive World Model Via Self-Supervised Learning}



\icmlsetsymbol{equal}{*}

\begin{icmlauthorlist}
\icmlauthor{Alana Santana}{ic}
\icmlauthor{Paula P. Costa}{feec}
\icmlauthor{Esther L. Colombini}{ic}
\end{icmlauthorlist}

\icmlaffiliation{ic}{Institute of Computing, State University of Campinas, Campinas, São Paulo, Brazil}
\icmlaffiliation{feec}{Faculty of Electrical and Computer Engineering, State University of Campinas, Campinas, São Paulo, Brazil}

\icmlcorrespondingauthor{Alana Santana}{alana.correia@ic.unicamp.br}

\icmlkeywords{Machine Learning, ICML}

\vskip 0.3in
]



\printAffiliationsAndNotice{\icmlEqualContribution} 

\begin{abstract}
Autonomous artificial agents must be able to learn behaviors in complex environments without humans to design tasks and rewards. Designing these functions for each environment is not feasible, thus, motivating the development of intrinsic reward functions. In this paper, we propose using several cognitive elements that have been neglected for a long time to build an internal world model for an intrinsically motivated agent. Our agent performs satisfactory iterations with the environment, learning complex behaviors without needing previously designed reward functions. We used 18 Atari games to evaluate what cognitive skills emerge in games that require reactive and deliberative behaviors. Our results show superior performance compared to the state-of-the-art in many test cases with dense and sparse rewards.
\end{abstract}

.

\section{Introduction}
\label{sec:introduction}



A prototypical Reinforcement Learning (RL) problem consists of an agent exploring the environment by choosing actions to maximize external rewards, also called extrinsic rewards. In many scenarios, these rewards are efficiently designed, but in the real world, they are scarce or non-existent. Nevertheless, autonomous artificial agents must be able to operate in complex environments without the need to pre-program rewards. This capability is still beyond the most advanced agents in the literature. In contrast, children exhibit a surprising ability to explore new environments, attend to objects, and physically engage with the outside world by creating new and exciting events \cite{haber2018learning}. Such behavior is considered a self-supervised learning process guided by internal motivations. Inspired by this phenomenon, RL theorists realized that other aspects than extrinsic reward must be considered for constructing intelligent agents.

\begin{figure}[htbp]
  \centering

   \includegraphics[width=0.5\linewidth]{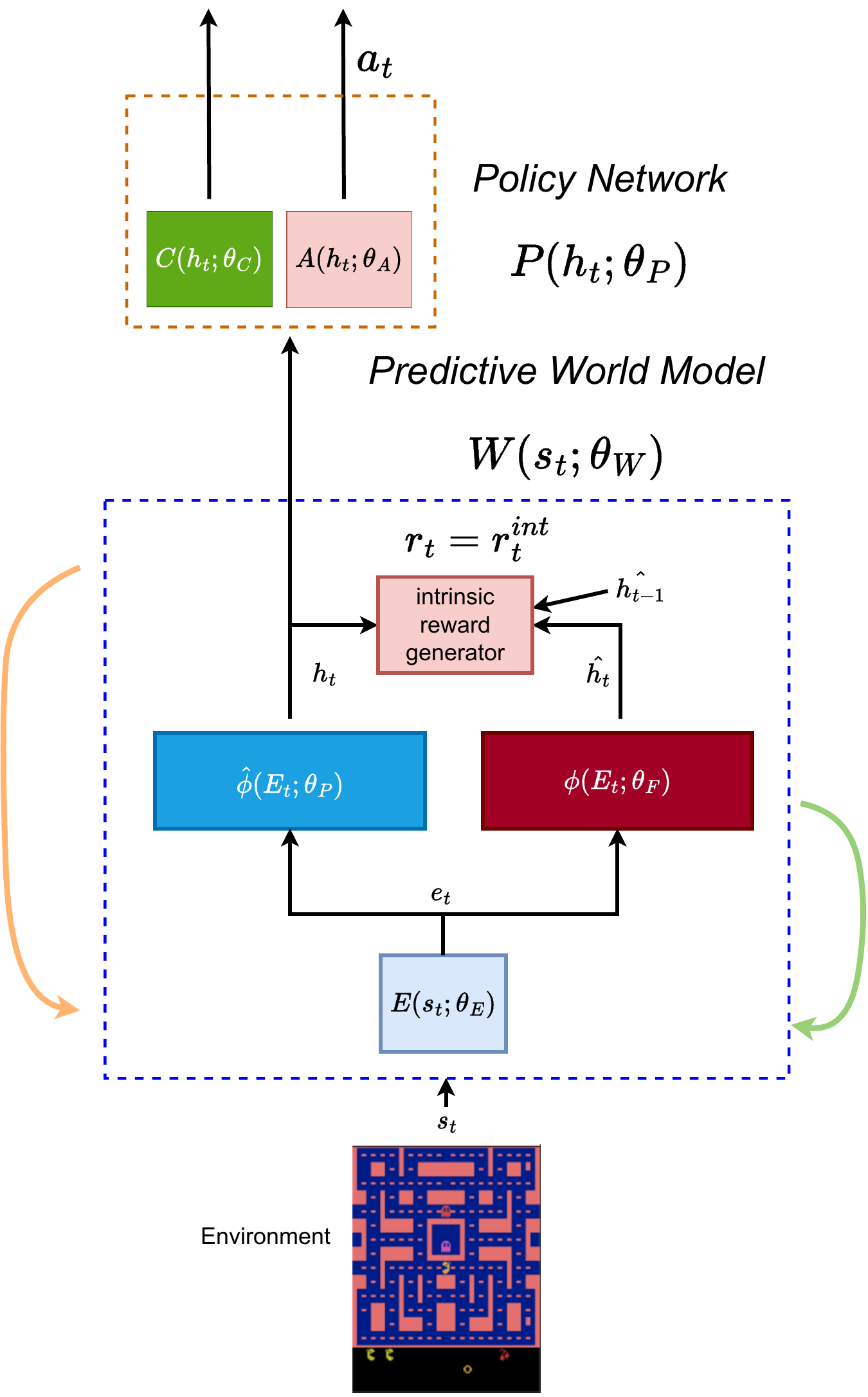}

   \caption{Our Intrinsically-motivated agent architecture. Our approach has two modules: predictive world model and policy network. The predictive world model generates intrinsic motivation rewards using attention and modular structures. At the same time, the policy network learns a policy to execute actions in the environment.}
   \label{fig:meu_modelo_simples}
\end{figure}

Evidence suggests that intrinsic motivation is vital to guide intellectual development \cite{barto2013intrinsic}. While extrinsic motivation implies an agent behaving to achieve goals by receiving external rewards, such as money or a prize, intrinsic motivation implies that the agent can reach goals through internal motivators, such as curiosity, enjoyment, and learning. The idea of intrinsic rewards originates in the work of Robert White \cite{white1959motivation} and D. E. Berlyne \cite{berlyne1966curiosity}. They criticize Freud’s work and Hullian’s view in which motivation is reduced only to drives related to biological needs (e.g., food). For example, the motivation for acquiring competence is not solely derived from energy sources conceptualized as drivers or instincts. Nevertheless, notions of novelty, complexity, and surprise are also drivers that motivate human behavior.

Intrinsic-motivated RL has led to state-of-the-art results in Agent57 \cite{badia2020agent57}: Deep Mind’s most successful Atari player. In Agent57, intrinsic motivation rewards provide more dense internal rewards for novelty-seeking behaviours, encouraging the agent to explore and visit as many states as possible. Despite significant advances in the field, many cognition elements still need to be addressed to build intrinsically motivated agents \cite{hao2023exploration}\cite{aubret2019survey}. According to Lecun \cite{lecun2022path}, we are equipped with an internal modular and hierarchical world that says what is probable, plausible, and impossible. Such a model, combined with simple behaviours and intrinsic motivations, guides the fast learning of new tasks, predicting the consequences of our actions, predicting the course of successful actions, and avoiding dangerous situations. However, according to him, building trainable internal models of the world that can deal with complex prediction uncertainties is still challenging because structural aspects of modularity and hierarchy have been neglected. 

The idea that humans use internal models of the world to learn has been around for a long time in psychology and neuroscience \cite{hawkins_framework_2019}\cite{hawkins_theory_2017}. Hawkins et al. \cite{hawkins_why_2017} worked studying human intelligence in the neocortex for at least two decades. They developed a simplified computational model of neocortical pyramidal cells and a local learning algorithm capable of updating the state of neurons and directing learning through non-expected events. Such a learning method updates the neuron weights only when the future state of sensorial receptive fields expectation is broken. According to him, the neocortex has a similar circuit composed of pyramidal cells, a highly modular and sparse hierarchical structure executing a common learning algorithm. Recently, Hole et al. \cite{hole_thousand_2021}, in a survey on the future of artificial intelligence, commented that to create truly intelligent agents, it is necessary to introduce essential aspects of the human neocortex, such as sparsity, independence, modularity, and hierarchy.

In this work, we propose to employ the cognition aspects of sparsity, modularity, independence, hierarchy, and attention to generate an internal world model for an intrinsically motivated agent. In this way, we guarantee the generation of the internal world with greater flexibility that better captures the physical aspects of the world. Specifically, our intrinsically-motivated agent has a predictive world model and a policy model. The predictive world model is entirely modular and hierarchical, composed of Bidirectional Recurrent Models \cite{mittal2020learning} that competitively generate representations of the agent's current and possible future state. Meanwhile, our policy network generates the agent's current action based on the current state. Figure \ref{fig:meu_modelo_simples} depicts our proposed approach, designed to support high dimensional data.

As our main contributions, we highlight the following:

\begin{enumerate}
    \item To the best of our knowledge, it is the first approach to adopt a predictive world model with sparsity, independence, modularity, and hierarchy aspects to create intrinsic rewards in RL agents.
    \item It combines modular attentional structures to improve the learning of intrinsic models.
    \item It allows advancing in learning models, achieving over 40\% of learning improvements in some test cases.
\end{enumerate}

\section{Related Work}
\label{sec:related_work}


Intrinsic motivation is a very studied topic in the reinforcement learning field, and a good summary is presented by Barto et al. \cite{barto2013intrinsic}, Aubret et al. \cite{aubret2019survey}, and Singh et al. \cite{singh2010intrinsically}. Initially, intrinsic motivation used concepts of emotion, surprise, empowerment, entropy, and information gain to formulate intrinsic rewards. Sequeira et al. \cite{sequeira2011emotion} explored the hypothesis that affective states encode information that guides an agent's decision-making during learning. Achiam et al. \cite{achiam2017surprise} proposed a surprise-based approach in which the agent learns a probability transition model of an MDP concurrently with the policy and generates intrinsic rewards that approximate the KL divergence of the learned model's true transition probabilities. Mohamed et al. \cite{mohamed2015variational} developed an approach using the empowerment concept. Variational autoencoders and convolutional neural networks produce a stochastic optimization algorithm directly from image pixels. Similarly, Klyubin et al. \cite{klyubin2005empowerment} used empowerment as the gain of information based on the entropy of actions to formulate intrinsic rewards.

Currently, approaches based on prediction error in the feature space have been extensively explored in the literature. In 2015, Stadie et al. \cite{stadie2015incentivizing} started their research using the feature space of an autoencoder to measure interesting states to explore. Pathak et al. \cite{pathak2017curiosity} proposed an approach based on an inverse dynamics model capable of scaling to high-dimensional continuous spaces and minimizing the difficulties of predicting directly in pixels, in addition to ignoring aspects of the environment that do not affect the agent. The approach showed that making predictions directly from the raw sensory space is unfeasible because it is challenging to predict pixels directly. Furthermore, some sensory spaces may be irrelevant to the agent's task. Agents trained with purely intrinsic rewards were able to learn task-relevant cognitive behaviors, demonstrating promising results in sparse environments. Similarly, Taylor et al. proposed an inverse dynamics model to assess the role of sensory space composition in the performance of an intrinsically motivated robotic arm that should manipulate objects on a table. Results showed that the approach works like an ``inside-out'' curriculum learning. The agent begins to explore its own body first, and only after acquiring knowledge does it explore its surroundings more frequently. Such results explain early motor behavior in infants and reinforce the hypothesis that discovering new patterns drives behavior.



Burda et al. \cite{burda2018large} investigated, in various Atari games, how curious agents and different feature spaces alter the results and performance of intrinsic agents. The results showed that: 1) generating the intrinsic reward from prediction error directly from the pixel space is challenging in high-dimensional environments; 2) variational autoencoders (VAEs) are a good summary of the observation but may contain many irrelevant details; 3) random features are fixed and insufficient in several scenarios; and 4) prediction error from inverse dynamic features is currently the best option to guarantee that the learned features contain essential aspects for the agent. Recently, Pathak et al. \cite{pathak2019self} presented an approach to deal with the challenge of stochasticity of environments. The authors used ideas from active learning to formulate an approach based on ensemble models.


Some experiments have shown that intrinsic rewards are indispensable for creating complex agents that supervise themselves in more realistic environments \cite{haber2018learning}. Haber et al. \cite{haber2018learning} demonstrated that the intrinsically motivated agents learned non-trivial cognitive behaviors such as self-generated motion (i.e., ego-motion), selective attention, and iteration with objects. The model presents two neural networks, the ``world-model'' which learns to predict the dynamic consequences of the agent's actions, while the ``self-model'' learns to predict errors in the agent's world model. The agent then uses the ``self-model'' to choose actions that it believes will adversely challenge the current state of its ``world-model''. This learning occurs through a self-supervised emergent process in which new abilities emerge in developmental milestones, as in human babies. In addition, the agent also learns improved visual encodings in specific tasks, such as detection, location, object recognition, and the prediction of physical dynamics better than other state-of-the-art approaches.


A key aspect that differentiates our work from others is that, to the best of our knowledge, it is the first approach to unite several elements of cognition that have been neglected for a long time. Based on studies proposed by Hawkins et al. \cite{hawkins_why_2017}, Lecun et al. \cite{lecun2022path}, and Hole et al. \cite{hole2021thousand}, we combine sparsity, modularity, hierarchy, and attention to building an internal world model for an intrinsically motivated agent. Our framework has much potential because it generates predictions of the agent's future states from a modular, hierarchical, and fully reconfigurable structure through bottom-up and top-down attentional signals. Such mechanisms allow the internal world model to generate future states' predictions from competitive small independent modules similar to the human neocortex.

\section{Bidirectional Recurrent Models}
\label{sec:preliminaries}

\textbf{Bidirectional Recurrent Models (BRIMs)} mainly use self-attention to link identical LSTM modules, generating a very sparse and modular framework with only a small portion of modules actives at time $t$~\cite{mittal2020learning}. The approach separates the hidden state into several modules so that upward iterations between bottom-up and top-down signals can be appropriately focused. The layer structure has concurrent modules so that each hierarchical layer can send information both in the bottom-up and top-down directions. Bottom-up attentional subsystems communicate between modules of the same layer, as well as the composition of hidden states in initial layers using the entry $x_{t}$ as the target, and via top-down attention modules in different layers communicate with each other requesting information about hidden states of previous and posterior layers to compose the current hidden state. BRIMs is composed of the following structures.

\textbf{Multi-layer Stacked Recurrent Networks}. Most multi-layer recurrent networks are configured to operate feed-forward and bottom-up, meaning that higher layers are fed with information processed by inferior layers. In this sense, the traditional stacked RNN for $L$ levels is defined as $\textbf{h}_t^l = F^l(\textbf{h}_t^{l-1}, \textbf{h}_{t-1}^l)$, where $l = 0, 1, ..., L$. For a specific time step $t$, $\textbf{y}_t = D(\textbf{h}_t^L)$ executes the prediction, based on input $\textbf{x}_t$, where $\textbf{h}_t^0 = E(\textbf{x}_t)$ is the first hidden state at model, and $\textbf{h}_t^l$ to the hidden state at layer $l$. $D$ defines the decoder, $E$ is the encoder, and $F^l$ represents the recurrent dynamic at layer $l$ (e.g., LSTM, GRU).





\textbf{Recurrent Independent Mechanisms (RIMs)}. Proposed by Goyal et al.~\cite{goyal2019recurrent}, RIM is a single-layered recurrent architecture that consists of hidden state $\textbf{h}_t$ decomposed into $n$ modules. The main property introduced in this model is that on a specific
time step, only a small subset of modules is activated. In this sense, the hidden states are updated following these steps: a) a subset of modules is activated depending on their relevance to the input; b) the activated modules independently process the information; c) the activated modules
have contextual information from the other modules and update their hidden state to store such information.

\textbf{Key-Value Attention}. The Key-Value Attention, also called the Scaled Dot Product, is responsible for the updates in RIM. This attentional mechanism is also employed in the self-attention modules, widely used in Transformer architectures. The attention score $\textbf{A}_S = \mathrm{Softmax} \bigg(\frac{\textbf{QK}^T}{\sqrt{d}}\bigg)$ and an attention modulated result $\textbf{A}_R = \textbf{A}_S\textbf{V}$ are computed by self-attention modules, where $\textbf{Q}$ is the set of queries, $\textbf{K}$ are the keys with $d$ dimensions and $\textbf{V}$ are the values.




\textbf{Selective Activation}. The selective activation is employed by defining that each module creates queries $\mathbf{\bar{Q}} = Q_{inp}(h_{t-1})$ which are then combined with the keys $\mathbf{\bar{K}} = K_{inp}(\phi, x_t)$ and values $\mathbf{\bar{V}} = V_{inp}(\phi, x_t)$ obtained from the input $x_t$ and zero vectors $\phi$ to get both
the attention score $\mathbf{\bar{A}}_S$ and attention modulated input $\mathbf{\bar{A}}_R$. Based on this attention score, a fixed number of modules $m$ are activated for which the input information is most relevant. In this sense, the null module provides no new information and has a low attention score. The activated set per time step is defined as $\mathcal{S}_t$.

\textbf{Independent Dynamics}. After the input is modulated by attention, each activated module has its hidden-state update procedure, as

\begin{equation}
\mathbf{\bar{h}}_{t,k} = \left\{\begin{NiceMatrix}[l]
F_k(\mathbf{\bar{A}}_{R_k}, \textbf{h}_{t-1,k}) & k \in \mathcal{S}_t
\\
\textbf{h}_{t-1,k} & k \notin \mathcal{S}_t,
\end{NiceMatrix}\right.
\end{equation}

\noindent where $F_k$ is any recurrent update procedure (e.g., GRU, LSTM).

\textbf{Communication}. Each module consolidates the information from all the other modules for every independent update step. The attention mechanism is utilized to consolidate this information in a similar way as in selective activation. The active modules create queries $\mathbf{\hat{Q}} = Q_{com}(h_t)$ which act with the keys $\mathbf{\hat{K}} = K_{com}(h_t)$ and values $\mathbf{\hat{V}} = V_{com}(h_t)$ generated by all modules and the result of attention $\mathbf{\hat{A}}_R$ is combined to the state in time step $t$ as

\begin{equation}
\textbf{h}_{t,k} = \left\{\begin{NiceMatrix}[l]
\mathbf{\bar{h}}_{t,k} + \mathbf{\hat{A}}_{R_{k}} & k \in \mathcal{S}_t
\\
\mathbf{\bar{h}}_{t,k} & k \notin \mathcal{S}_t.
\end{NiceMatrix}\right.
\end{equation}

\textbf{Composition of Modules}. The original hidden state $\mathbf{h}^l_t$ found in RIM is decomposed for each layer $l$ and time $t$ into separate modules. Therefore, instead of representing the state as just a fixed dimensional vector $\mathbf{h}^l_t$, the representation is defined as $\{((\mathbf{h}^l_{t,k})^{n_l}_{k=1}, \mathcal{S}^l_t)\}$ where $n_l$ denotes the number of modules in layer $l$ and $\mathcal{}S^l_t$ is the set of active modules at time $t$ in layer $l$. $|\mathcal{S}^l_t| = m_l$, where $m_l$ is a hyperparameter to define the number of modules active in each layer $l$ at any time; layers may have a different number of active modules. Setting $m_l$ to be half of $n_l$ provided good performance.
 
\textbf{Communication Between Layers}. The communication links are defined between
multiple layers using key-value attention. Tradicional RNNs build a strictly bottom-up multi-layer dependency; in BRIMs, instead, the multi-layer dependency considers queries $\mathbf{\bar{Q}} = Q_{lay}(\mathbf{h}^l_{t-1})$ from modules in layer $l$, and keys $\mathbf{\bar{K}} = K_{lay}(\phi, \mathbf{h}^{l-1}_t, \mathbf{h}^{l+1}_{t-1})$ and values $\mathbf{\bar{V}} = V_{lay}(\mathbf{\phi}, \mathbf{h}^{l-1}_t, \mathbf{h}^{l+1}_{t-1})$ from all the modules in the lower and higher layers. Thus, the attention mechanism acts in three directions and generates the attention score $\mathbf{\bar{A}}^l_S$ and output $\mathbf{\bar{A}}_R$. The same layer gives the attention-receiving information from the higher layer in the previous time step; the same layer also gives the attention-receiving information from the lower layer in the current time step. Only the lower layer is used for the deepest layer, and for the first layer, the input’s embedded state serves as the lower layer~\cite{mittal2020learning}.

\textbf{Sparse Activation}. The set $S^l_t$ is built based on the attention score $\mathbf{\bar{A}}^l_S$, which contains modules for which null information has little importance. Every activated module gets a separate input version, which is obtained via the attention output $\mathbf{\bar{A}}^l_R$. In practice, for each activated module, the representation is defined as $\mathbf{\bar{h}}_{t,k}^{l} = F_k^l(\mathbf{\bar{A}}_{R_k}^l, \textbf{h}_{t-1,k}^l)$, where $k \in \mathcal{S}_t^l$, and $F^l_k$ represents the recurrent update unit.



\textbf{Communication Within Layers}. Communication between the different modules within each layer using the key-value attention. This communication between modules within a layer permits the modules to share information through the bottleneck of attention. In the same way, queries are generated $\mathbf{\hat{Q}} = Q_{com}(\mathbf{\bar{h}}^l_t)$ from active modules and keys $\mathbf{\hat{K}} = K_{com}(\mathbf{\bar{h}}^l_t)$ and values $\mathbf{\hat{V}} = V_{com}(\mathbf{\bar{h}}^l_t)$ from all the modules to obtain the final update to the module state through residual attention $\mathbf{\hat{A}}^l_R$. The state update rule is

\begin{equation}
\textbf{h}_{t,k}^l = \left\{\begin{NiceMatrix}[l]
\mathbf{\bar{h}}_{t,k}^{l} + \mathbf{\bar{A}}_{R_{k}}^l & k \in \mathcal{S}_t^l
\\\mathbf{\bar{h}}_{t-1,k}^l & k \notin \mathcal{S}_t^l.
\end{NiceMatrix}\right.
\end{equation}


\section{Proposed Model}
\label{sec:method}

\begin{figure*}[htb]
  \centering

   \includegraphics[width=0.65\linewidth]{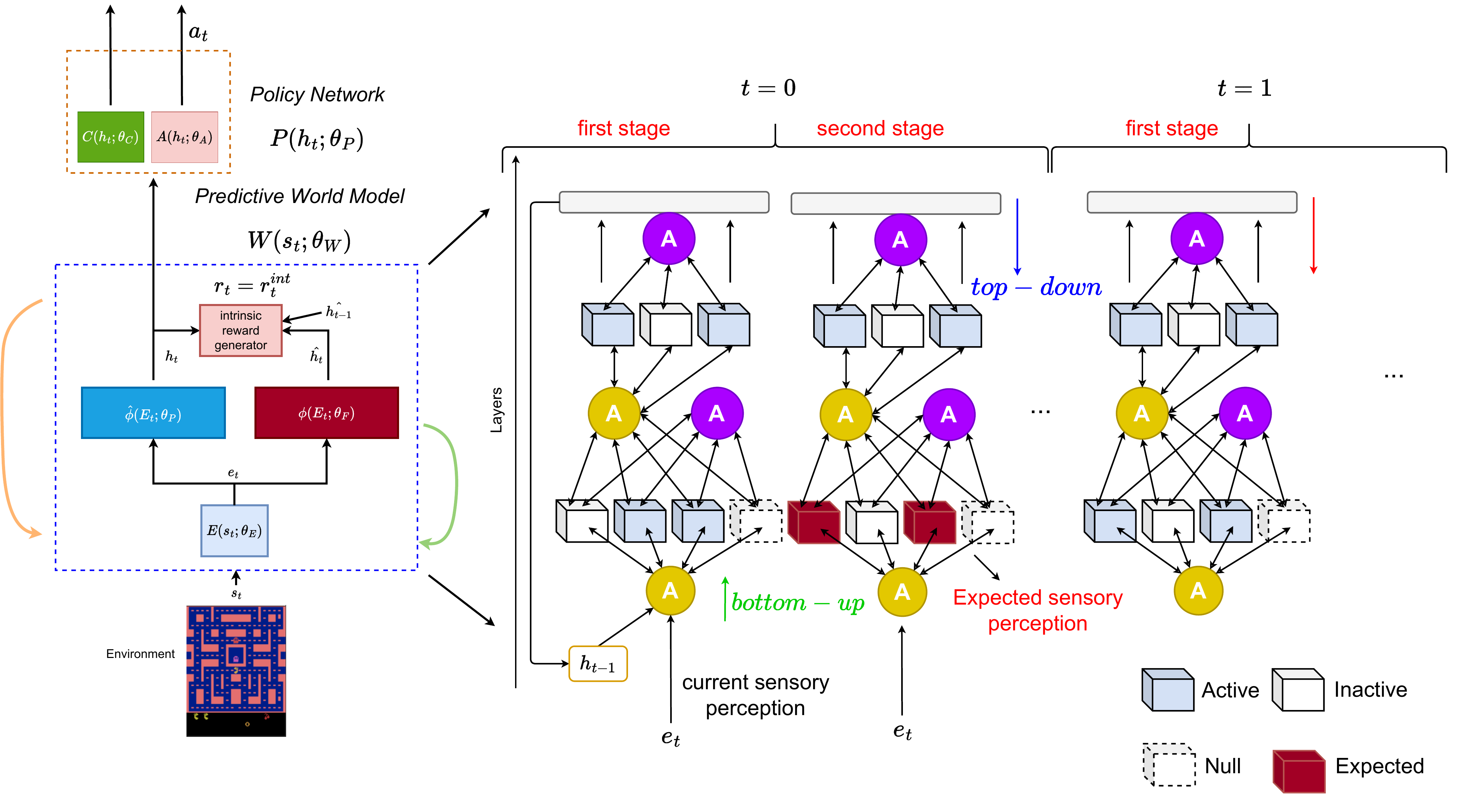}

   \caption{Intrinsically-motivated agent architecture. At each time step $t$, the current state $s_{t}$ triggers the predictive world modules. Modules can be active, expected, null, or inactive state. Active modules use the state information $s_{t}$ to build a representation for choosing the current action. Modules in the expected state build an expected representation for the next state $s_{t+1}$ before the agent sees it. Thus, the agent performs a future prediction of the consequences of its action in the world. Finally, inactive/null modules do not participate in the gradient and can be activated as needed in the next iteration. After executing the action, the intrinsic reward is the difference between the agent's expectations and the real-world state.}
   \label{fig:model_complete}
\end{figure*}

Our intrinsically-motivated agent has two models: 1) \textbf{the predictive world model} composed of a modular and competitive structure to generate the intrinsic rewards; 2) \textbf{the policy model} that learns a policy capable of generating a sequence of actions to maximize the reward signal. At the end of each action performed by the agent in the environment, it receives an intrinsic reward $r^{int}_{t}$ generated by the predictive world model, as shown in Figure \ref{fig:model_complete}.

The predictive world model $W$ is the key to our development. It is represented by a set of neural networks parameterized by $\theta_{E}$, $\theta_{P}$, and $\theta_{F}$. The model comprises a feature encoding $E$ and independent recurrent modules that competitively generate representations of the agent's current and possible future state. At each time step $t$, the encoder $E$ receives the current state $s_{t}$ from the agent and generates a vector $x_{t}$ that encodes all the visual information. We used three 2D convolutional layers with linear layers to perform the encoding in our case. A key-value attention receives $x_{t}$ as a query and determines which independent recurrent modules should enter an active, inactive, or expected state. In summary:

\begin{equation}
  x_{t} = E\left ( s_{t} ; \theta_{E} \right ),
  \label{eq:formule_1}
\end{equation}

\begin{equation}
  h_{t}^{l} = \phi \left ( x_{t}, h_{t-1}^{l} ; \theta_{P} \right ),
  \label{eq:formule_2}
\end{equation}

\begin{equation}
  \hat{h_{t}^{l}} = \hat{\phi} \left ( x_{t}, \hat{h^{l}}_{t-1} ; \theta_{F} \right ),
  \label{eq:formule_3}
\end{equation}


\noindent where $h_{t}^{l} = \left \{ h^{l}_{t, 1}, h^{l}_{t, 2}, ..., h^{l}_{ t, k}\right \}$, $k \in S_{t}^{p}$ is an embedding vector composed by activated modules with the current state $s_{t}$. $S_{t}^{ p}$ are indices of activated modules, and $\hat{h_{t}^{l}} = \left \{ \hat{h^{l}} _{t, 1}, \hat{h^{l}}_{t, 2}, ..., \hat{h^{l}}_{t, w}\right \}$, with $w \in S_{t}^{f}$, is an embedding vector composed by modules triggered in the expected state. $S_{t}^{f}$ are indices of modules in the expected state. $\phi$ and $\hat{\phi}$ are the LSTMs representing the modules.


The active modules use the encoder's information, passing it to the successive layers. At each time step $t$, when a module is inactive, it does not incorporate the new information, and there is no gradient flow. In parallel, some modules fire in the expected state (i.e., the modules activated to generate a future prediction to the next state $s_{t+1}$). As the structure is modular, the world representation is divided among the different modules; together, they make up a complete representation of the world. The attentional bottleneck directs the different modules' activation/deactivation/expectation flow so that the agent's internal representations of the world are useful for its actions without the need for inverse dynamics models. In addition, each module becomes an expert in certain aspects of the environment.

Attention-driven bottom-up and top-down signals are especially beneficial in selecting agent actions. In phases where bottom-up information dominates, it can benefit the agent to act immediately to resolve unforeseen events. At the same time, top-down signals can be useful when the system needs a long-term plan. Furthermore, this representation is very similar to the activation and deactivation of pyramidal cells in the human neocortex \cite{hawkins2017theory}.


Overall, attention-guided dynamic activation is essential in intrinsic agent learning, given that there is evidence in the literature that the modular structure can better handle distribution changes implicit in training as the policy evolves. At the end of the competition between modules, two state representations are created, $h_{t}^{l}$ and $\hat{h_{t}^{l}}$, respectively. $h_{t}^{l}$ is the latent space vector of the agent's current state and will be input to the policy network. In contrast, $\hat{h_{t}^{l}}$ is the representation expected to the next state $s_{t+1}$ after the agent executes the current action. Learning occurs when an expectation break is performed between $h_{t}^{l}$ and $\hat{h_{t}^{l}}$, so that the intrinsic reward is generated by $r^{int}_{t} = \frac{\left \| h_{t}^{p} - h_{t-1}^{f} \right \|^{2}_{2}}{n}$, where $n$ is the size of the vector. Finally, we represent the policy $\pi \left ( h_{t}^{p}; \theta_{P} \right )$ by a deep neural network with parameters $\theta_{P}$. Given the agent in state $s_{t}$, it executes action $a \sim \pi \left ( h_{t}^{p}; \theta_{P} \right ) $ sampled from the policy. $\theta_{E}$, $\theta_{P}$, and $\theta_{G}$ are optimized to maximize the expected sum of intrinsic rewards, given by




\begin{equation}
  \underset{\theta_{E}, \theta_{P}, \theta_{G}}{\textup{max}} E_{\pi \left ( h_{t}^{p}; \theta_{P} \right )}\left [ \sum_{t} r_{t}^{int}\right ].
  \label{eq:max_rewards}
\end{equation}


In parallel, $\theta_{E}$ and $\theta_{F}$ are minimized by a regression loss, given by

\begin{equation}
  \underset{\theta_{E}, \theta_{F}}{\textup{min}} L_{W}(h_{t}, \hat{h}_{t-1}).
  \label{eq:max_rewards_2}
\end{equation}

\noindent where $L_{W}$ measures the discrepancy between the predicted and actual features and is modelled as the mean squared error function. 

The overall optimization problem can be written as

\begin{equation}
  \underset{\theta_{E}, \theta_{P}, \theta_{F}, \theta_{G}}{\textup{min}} \left [ E_{\pi \left ( h_{t}^{p}; \theta_{P} \right )}\left [ \sum_{t} r_{t}^{int}\right ] + L_{W} \right ].
  \label{eq:max_rewards_2}
\end{equation}

We propagate the gradient of the policy and world models through the encoder to ensure that the generated representation is useful for the agent's task.


\section{Experiments}
\label{sec:experiment}

In this section, we conduct a set of experiments to evaluate our proposed method.

\subsection{Experiment Setup}
\label{experiment_setup}

\textbf{Environments.} We evaluate our method on 18 Atari Games, a standard setup for evaluating Reinforcement Learning methods. We choose Atari games with varying reward types (i.e., dense, sparse, and hybrid) and cognitive skills that the agent needs to learn. We aim to evaluate the agent in different games to analyze our approach's positive and negative points. Some games chosen are dynamic and require the agent to show fast and efficient reactive behaviors to survive in the environment. In contrast, other games are strategy games and require the agent to think of a plan to achieve a specific goal. 


\textbf{Architecture and pre-processing.} We used the PPO (Proximal Policy Optimization) algorithm \cite{schulman2017proximal}, a robust learning algorithm that requires little hyperparameter tuning, for our experiments. While training with PPO, we normalize the advantages \cite{sutton2018reinforcement} in a batch to have a mean of 0 and a standard deviation of 1. We use the same architecture to train all games. We use one layer of Recurrent Independent Mechanisms in the predictive model with eight modules, only 4 of which can be on active/expected states in each time step $t$. Two modules represent the current state $s_{t}$, two generate the expected representation to $s_{t+1}$, and four are inactive. Each module has a size of 32. All experiments were carried out in the pixels space. We converted all images to grayscale and resized them to 84 $\times$ 84. We represent the state as a stack of historical observations $\left [ x_{t-3}, x_{t-2}, x_{t-1}, x_{t}\right ]$ to deal with partial observability. We also use the standard 4-frame frameskip. Unlike the classical methods, we perform a simple state normalization by dividing by 255.

\textbf{Hyperparameters and implementation details.} We used a learning rate of 0.00025 for all networks. In all experiments, we used 128 parallel environments, rollouts of length 128, and four epochs. During training, we limited the maximum episode size to 4,500 steps to avoid the collapse of the learned policy. We train most models with 170 million steps. However, we noticed that our model quickly surpassed the state of the art, and we reduced some training steps to 50 million. We consider death endgame in all experiments.

\textbf{Evaluations.} To evaluate the agents' performance, we use the extrinsic reward, which is the game's score at the end of each episode. After training, we put the agents to play ten games with different seeds from those used during training. Finally, we computed the mean and standard deviation of the game scores obtained.

\textbf{Hardware and Software Configurations.} We implemented our method with PyTorch 1.3.1 and CUDA v11.1. We conducted experiments on Nvidia RTX 8000 with 48Gb, motherboard Asus Rog Strix Z490-E Gaming, Intel(R) Xeon(R) Gold 6230 CPU @ 2.10GHz, RAM Corsair DDR4 125Gb Gb @ 3600MHz, disk Western Digital 1Tb, and Operating System Ubuntu 20.04.3 LTS.

\subsection{Results}

In this section, we present the results obtained from our experiments. Our code is publicly available in our repository\footnote[1]{https://github.com/XXXXX/XXXXX}. We compare our results with those obtained by the model proposed by Pathak et al. \cite{pathak2017curiosity}, which was further explored by Burda et al. \cite{burda2018large} using Atari games. We chose this model for comparison among many other works because it is a state-of-art approach to comparing results. In addition, it presents an extensive study of Atari games and is what most relates to our method. Our experiments showed that the observation normalization process adopted in \cite{burda2018large} typically resulted in policy collapse after 50 million training steps. While this is an undesired behavior, we adopted the same normalization strategy (Section \ref{experiment_setup}) for comparison purposes. We trained 18 Atari games with different gameplay features and reward patterns. All results are shown in Table \ref{tab:tabela_resultados}. The results show that we scored quite significant games on $90\%$ on cases of the test.

\begin{table}[htb]
\centering
\begin{tabular}{ccccc}
\hline
Environment       & Ours                  & Burda \\ \hline
MsPacman     &  1358.0 $\pm$ 439.7 & 380.0 $\pm$ 113.3   \\ 
Atlantis   &         47540.0 $\pm$ 7804.1               &   9800.0 $\pm$ 4475.2                          \\
Freeway    &             3.2 $\pm$ 1.3                &     1.0 $\pm$ 0.29                               \\
Asterix    &           2465.0 $\pm$ 1235.9               &      110. $\pm$ 48.9                        \\
RiverRaid  &             4123.0 $\pm$ 1837.8               &    798.0 $\pm$ 299.1                        \\
Asteroids  &            821.0 $\pm$ 273.7              &        543.0 $\pm$ 192.6                          \\
Jamesbond  &                245.0 $\pm$ 89.0               &        35.0 $\pm$ 19.0                           \\
Centipede  &                6128.0 $\pm$ 2709.0               &      1620.9 $\pm$ 983.2                          \\
KungFuMaster     &              510.0 $\pm$ 361.8               &     80.0 $\pm$ 10.8                         \\
Solaris    &                 1658.0 $\pm$ 564.0               &       2190.0 $\pm$ 562.0                 \\
Pitfall    &              -366.9 $\pm$ 317.3               &        -512.2 $\pm$ 489.9                             \\
Alien      &                536.0 $\pm$ 150.0               &         291 $\pm$ 83.8                            \\
Robotank   &               7.8 $\pm$ 3.9                 &           2.6 $\pm$ 1.2                    \\
BeamRider  &            988.0 $\pm$ 437.8                &          347.6 $\pm$ 123.5                              \\
Amidar     &            65.8 $\pm$ 15.6                 &          14.1 $\pm$ 3.1                        \\
BattleZone &            6300.0 $\pm$ 4838.3                 &   1900.0 $\pm$ 577.                                     \\
Seaquest   &            402.0 $\pm$ 60.9                   &         88.0 $\pm$ 14.4                           \\
Gravitar   &            215.0 $\pm$ 184.4                 &        280 $\pm$ 106.0                      \\ \hline
\end{tabular}
\caption{Our results on the Atari games. We used, as a result, the mean and standard deviation of scores received in ten games. Our approach outperforms the baseline in several games, especially MsPacman, Atlantis, Asterix, and BattleZone.}
\label{tab:tabela_resultados}
\end{table}

\begin{figure}[htb]
     \centering
         \centering
         \includegraphics[width=7.5cm, height=4.0cm]{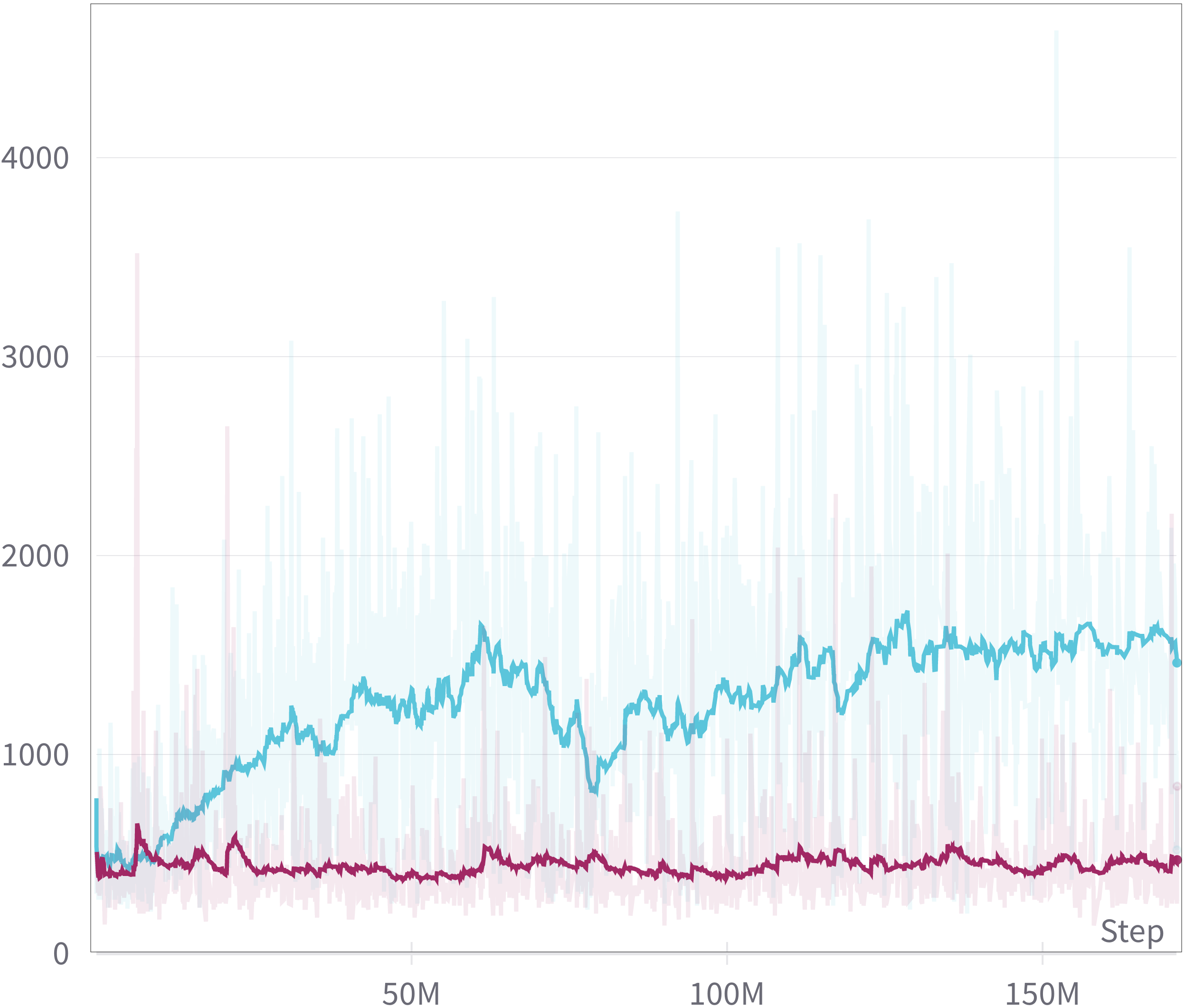}
         \caption{MsPacman game. The training curve shows the best extrinsic returns per episode in the MsPacman environment. In blue, we have our agent, and in red, we have the baseline. The x-axis represents the training steps, and the y-axis is the game score (i.e., accumulated extrinsic reward) received at the end of the episode.}
        \label{fig:pacman_resultados}
\end{figure}

\textbf{Reactive and deliberative environments}. These environments require different cognitive skills from the agent over time, and the target distribution changes implicitly throughout training. In our experiments, MsPacman, Alien, and Amidar are environments that require reactive and deliberative behaviors. Our agent explored various policies and learned faster, well-aligned behaviors without extrinsic rewards. Figure \ref{fig:pacman_resultados} shows that our agent quickly learned to iterate in the MsPacman environment, achieving a learning curve much higher than the baseline. MsPacman is an interesting game to analyze. Initially, the rewards are dense, and the agent needs to be reactive to escape the ghosts. As the agent advances in the environment, the game gradually becomes sparse, demanding more elaborate planning strategies. The results show that through the modular structure, our agent can better capture the world's compositional structure and have a concise and efficient representation to generate its actions. Attention and modular structures demonstrate greater flexibility to generate low-dimensional features, efficiently filtering out irrelevant parts of the observation space. The modular structures also demonstrate efficiency since the agent seems to have all the essential information to perform the actions, achieving an average performance of $50\%$ higher than the baseline during learning. Also, our approach deals better with non-stationary rewards, as we do not provide any standardization to intrinsic rewards.

\begin{figure*}[htb]%
    \centering
    \subfigure[\centering Asterix]{{\includegraphics[width=5cm]{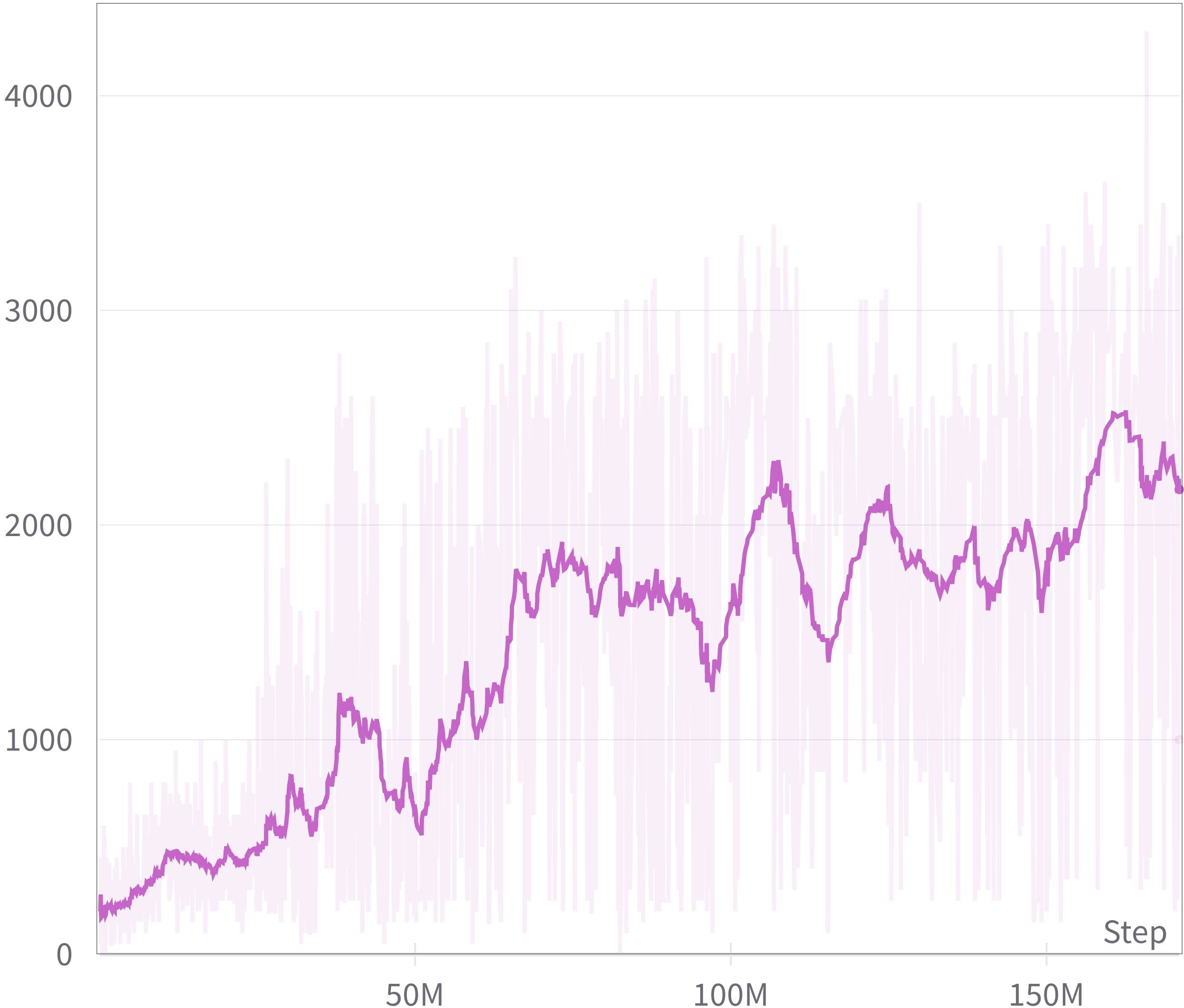} }}%
    \qquad
    \subfigure[\centering RiverRaid]{{\includegraphics[width=5cm]{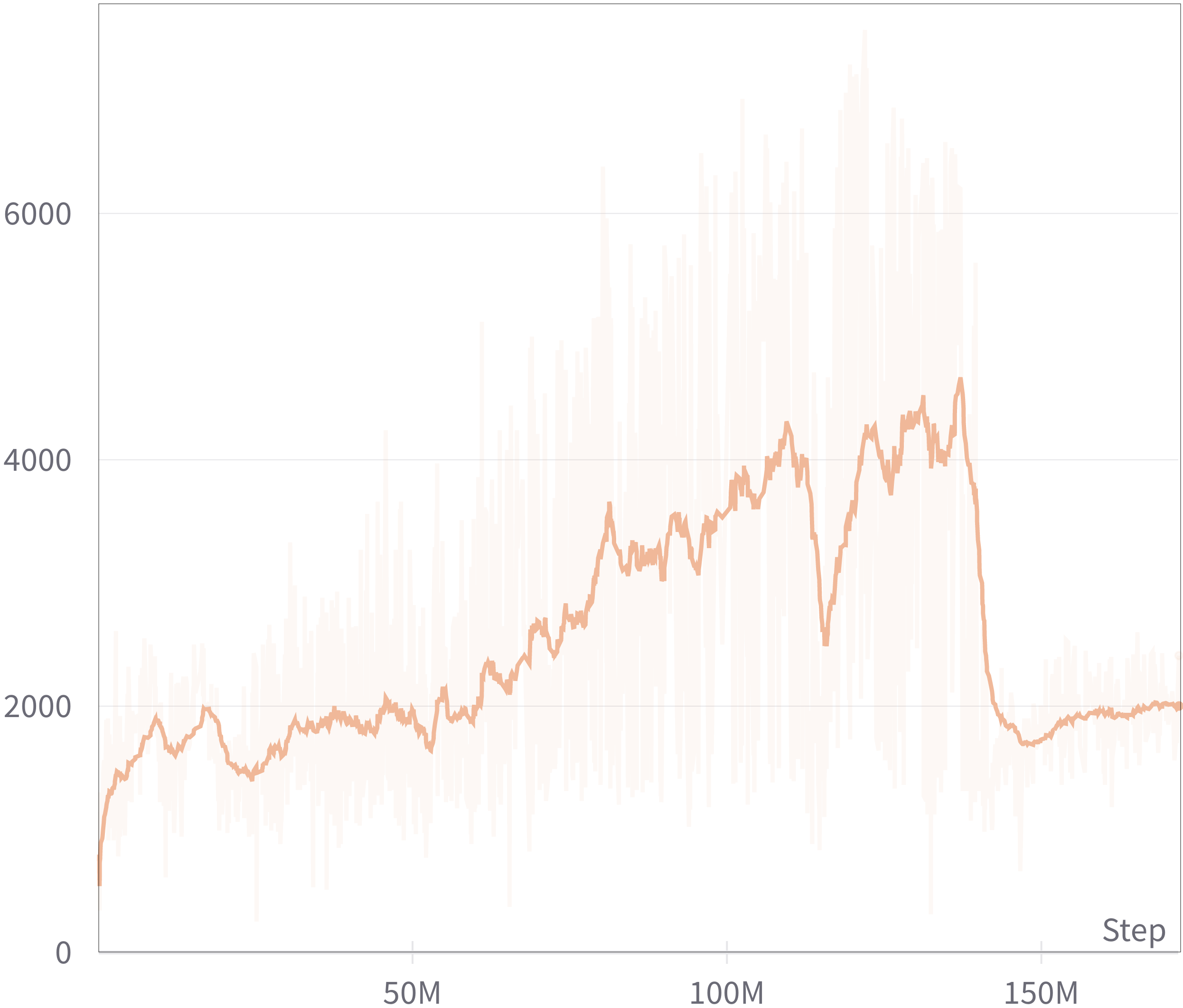} }}%
    \qquad
    \subfigure[\centering Atlantis]{{\includegraphics[width=5cm]{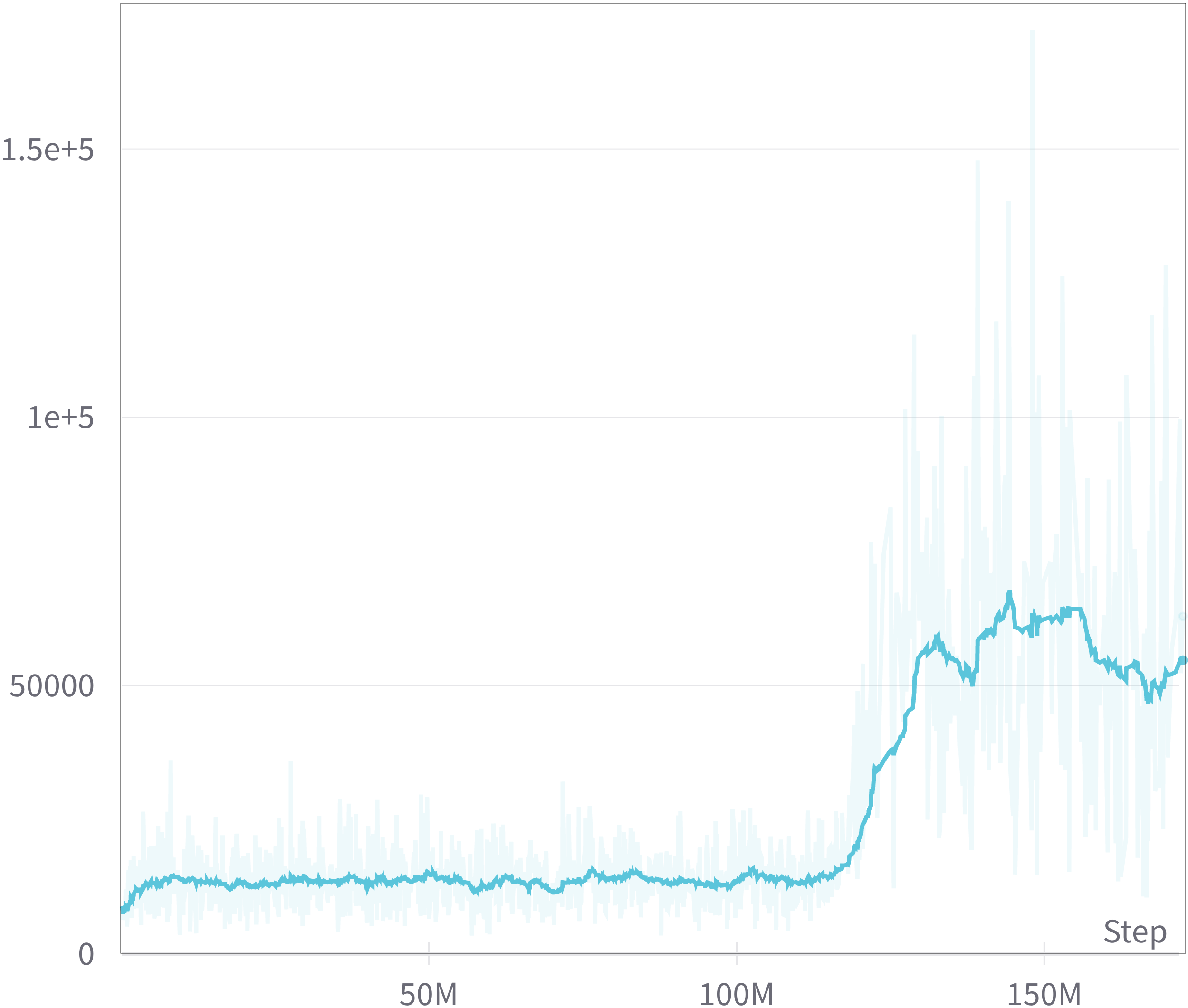} }}%
    \caption{Our proposed approach results. The training curve shows the best extrinsic returns per episode in the Asterix, Riverraid, and Atlantis environments. The x-axis represents the training steps, and the y-axis is the game score (i.e., accumulated extrinsic reward) received at the end of the episode. In these scenarios, our agent excels against the baseline (Table \ref{tab:tabela_resultados}). The exploratory strategies chosen to stay alive allowed the agent to obtain a much higher score in these games. After exploring states that led to bad scores, the agent quickly changes its exploratory strategy.}%
    \label{fig:todos_resultados}%
\end{figure*}

\textbf{Purely reactive environments.} In purely reactive environments, our agent outperforms the baseline, as seen in the Asterix, Centipede, and Riverraid games. In these environments, the agent must interact quickly with enemies to stay alive. This type of game is exciting in intrinsic motivation because as the agent interacts with other agents, patterns more challenging to predict eventually emerge. As many novelties arise, the agent remains motivated to explore new behaviors for extended periods. The agent notices that death leads to an early ending of the episode, and fighting to stay alive allows it to discover new states that have not yet been explored. In this scenario, our agent stands out against the baseline. The exploratory strategies chosen to stay alive allowed the agent to achieve a much higher score in these games. Figure \ref{fig:todos_resultados} a) and b) clearly shows that the agent continually learns coherent cognitive behaviors in the game that correlate well with the extrinsic reward. We believe this result is due to attention-guided representations, allowing specific modules to focus on enemies and forget about aspects of the environment that are irrelevant to staying alive. Consequently, the agent makes choices that score more in the game.

\begin{figure}[htb]
     \centering
     \includegraphics[width=7.5cm, height=4.0cm]{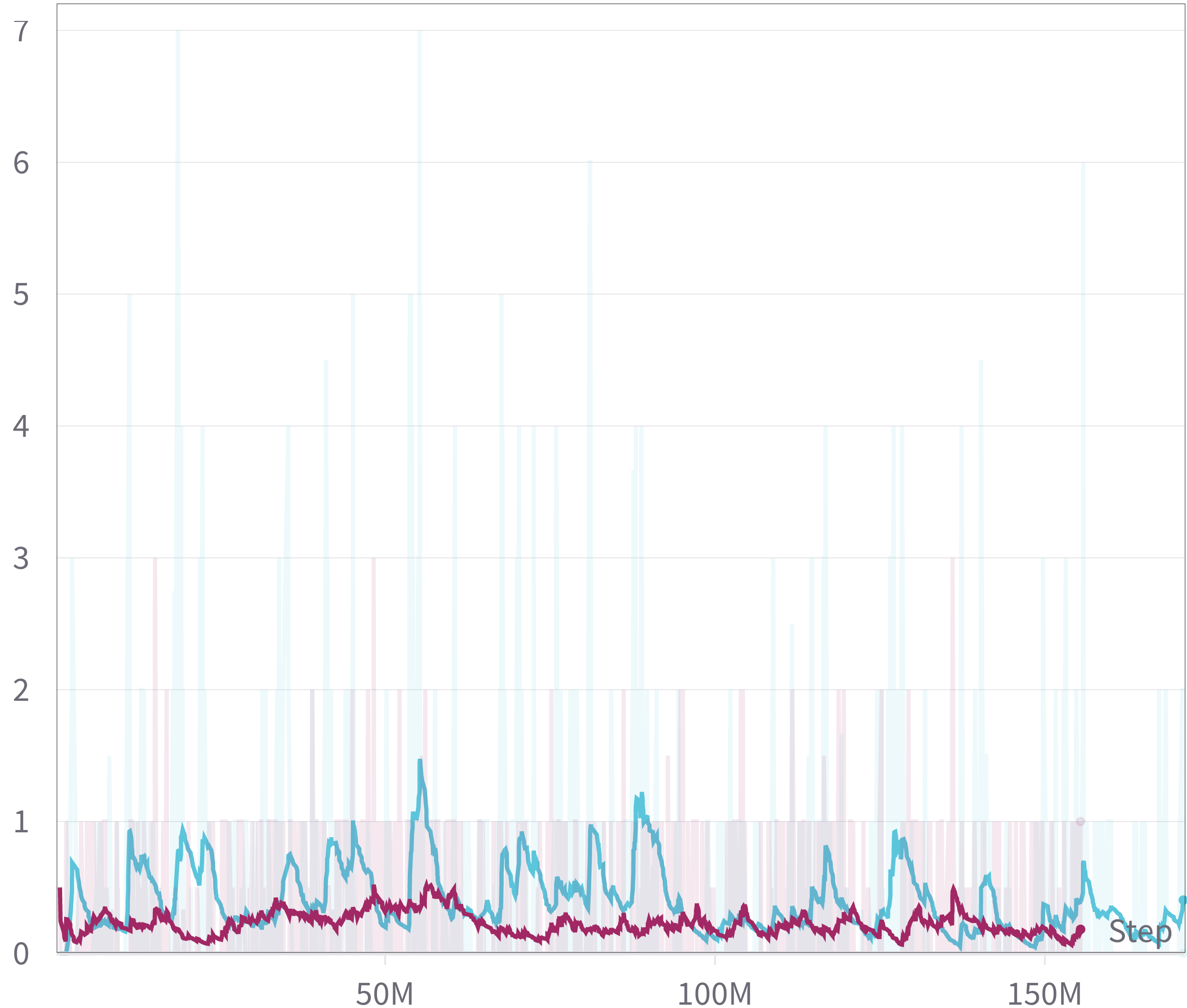}
     \caption{Freeway game. The training curve shows the best extrinsic returns per episode in the Freeway environment. In blue, we have our agent, and in red, we have the baseline. The x-axis represents the training steps, and the y-axis is the game score (i.e., accumulated extrinsic reward) received at the end of the episode.}
    \label{fig:freeway_resultados}
\end{figure}

\textbf{Sparse Rewards.} Significant results are also seen in purely sparse environments requiring agent planning. Figure \ref{fig:freeway_resultados} shows that our agent plans better in the Freeway environment and crosses the street more than the baseline. Sometimes, our agent can perform seven crossings on the street, while the baseline achieves a maximum of 3. Executing seven turns in the Freeway environment is relatively challenging since, at each crossing step the agent takes, an enemy can run over it. Only with a well-defined planning strategy can the agent perform such an action. We also overcame the baseline in the Pitfall environment. However, we did not get a positive score in any test cases. We believe this negative rating is due to the size of the rollout chosen for this environment. The 128 rollout is relatively small for sparse environments.

\textbf{Efficient exploration.} The intrinsic agents can fail to explore efficiently. In some situations, they do not learn highly correlated behaviors with extrinsic rewards. All our results show that our agent can more efficiently explore the environment than the baseline and learn more correlated behaviors with extrinsic rewards. Atlantis and Alien are games in which we obtained surprising results. The baseline cannot surpass a random agent that scores an average of 10,000 points in Atlantis and an average of 200 points in Alien, as shown in \cite{burda2018large}. Initially, our agent starts training by adopting a random exploratory similar to the baseline. However, as training progress, our agent chooses policies highly correlated with the extrinsic rewards, as shown in Figure \ref{fig:todos_resultados}c). These results demonstrate that our agent can choose more efficient exploratory strategies that lead to structured cognitive behaviors to perform a task even without any extrinsic reward.
\section{Conclusion}
\label{sec:conclusion}



In this work, we present a new approach to generating intrinsic rewards. We have demonstrated that our purely intrinsic attentional and modular agent can play multiple Atari games without external rewards. Our agent learns to perform complex behaviors faster during training. Our approach is simpler, demonstrating a greater alignment of the agent's actions with the extrinsic rewards given by the environment. 


As a result, we obtained results superior to the state-of-the-art in $90\%$ of the tested games. Our approach has had surprising results in highly dynamic environments, such as Atlantis, Asterix, Assault, and Riverraid, that require highly reactive agent behaviors. The attention and modular structure contributed significantly to a concise and efficient world representation, providing the agent with only what is essential for the current action. However, our approach has limited performance in highly sparse environments such as Pitfall. This result is due to the low number of rollouts used in the experiments for this environment. Future work will address how cognitive behaviours emerge in complex humanoids to object manipulation tasks in highly sparse and realistic environments.

\section*{Acknowledgements}
We thank CAPES, Quinto Andar, and H.IIAC for their financial support throughout the development of this work.
This study was financed in part by the Coordenação de Aperfeiçoamento de Pessoal de Nível Superior – Brasil (CAPES) – Finance Code 001. This project was supported by the Brazilian Ministry of Science, Technology and Innovations, with resources from Law nº 8,248, of October 23, 1991, within the scope of PPI-SOFTEX, coordinated by Softex and published Apredizado em Arquiteturas Cognitivas (Phase 3), DOU 01245.003479/2024-10.


\bibliography{example_paper}
\bibliographystyle{icml2023}



\end{document}